\documentclass[sigconf]{acmart2}

\usepackage{multirow}
\usepackage[capitalize]{cleveref}
\crefname{section}{Sec.}{Secs.}
\Crefname{section}{Section}{Sections}
\Crefname{table}{Table}{Tables}
\crefname{table}{Tab.}{Tabs.}

\def\eg{\textit{e.g.}}
\def\ie{\textit{i.e.}}

\def\eta{\textit{et al.}}

\newcommand{\myparagraph}[1]{{\setlength{\parskip}{0.3em} \noindent \textbf {#1}}}

\usepackage{xcolor}       %
\usepackage{hyperref}     %

\hypersetup{
    colorlinks=true,
    linkcolor=magenta,
    urlcolor=magenta,
    citecolor=magenta
}

\AtBeginDocument{%
  }

\setcopyright{acmlicensed}
\copyrightyear{2025}
\acmYear{2025}
\acmDOI{XXXXXXX.XXXXXXX}
\acmConference[Conference'25]{xxx}{2025}{xxx}

\acmISBN{978-1-4503-XXXX-X/2018/06}

\begin{document}

\title{Let Your Video Listen to Your Music! -- Beat-Aligned, Content-Preserving Video Editing with Arbitrary Music}

\author{Xinyu Zhang\textsuperscript{1}, Dong Gong\textsuperscript{2}, Zicheng Duan\textsuperscript{1}, Anton van den Hengel\textsuperscript{1}, Lingqiao Liu\textsuperscript{1}}
\affiliation{%
\institution{\textsuperscript{1}The University of Adelaide\quad \textsuperscript{2}University of New South Wales}  
  \country{Project page: \color{magenta}\href{https://zhangxinyu-xyz.github.io/MVAA/}{ $\tt zhangxinyu$-$\tt xyz.github.io/MVAA$}}
}

\begin{abstract}

Aligning the rhythm of visual motion in a video with a given music track is a practical need in multimedia production, yet remains an underexplored task in autonomous video editing. Effective alignment between motion and musical beats enhances viewer engagement and visual appeal, particularly in music videos, promotional content, and cinematic editing. Existing methods typically depend on labor-intensive manual cutting, speed adjustments, or heuristic-based editing techniques to achieve synchronization. While some generative models handle joint video and music generation, they often entangle the two modalities, limiting flexibility in aligning video to music beats while preserving the full visual content. In this paper, we propose a novel and efficient framework—termed MVAA (Music-Video Auto-Alignment)—that automatically edits video to align with the rhythm of a given music track while preserving the original visual content. To enhance flexibility, we modularize the task into a two-step process in our MVAA: aligning motion keyframes with audio beats, followed by rhythm-aware video inpainting. Specifically, we first insert keyframes at timestamps aligned with musical beats, then use a frame-conditioned diffusion model to generate coherent intermediate frames, preserving the original video’s semantic content. Since comprehensive test-time training can be time-consuming, we adopt a two-stage strategy: pretraining the inpainting module on a small video set to learn general motion priors, followed by rapid inference-time fine-tuning for video-specific adaptation. This hybrid approach enables adaptation within ~10 minutes with one epoch on a single NVIDIA 4090 GPU using CogVideoX-5b-I2V~\cite{yang2024cogvideox} as the backbone. Extensive experiments show that our approach can achieve high-quality beat alignment and visual smoothness. User studies further validate the natural rhythmic quality of the results, confirming their effectiveness for practical music-video editing.

\end{abstract}

\ccsdesc[500]{Computing methodologies~Computer vision}
\keywords{Video-Music alignment; Video editing; Diffusion model}
\begin{teaserfigure}
  \includegraphics[width=0.95\textwidth]{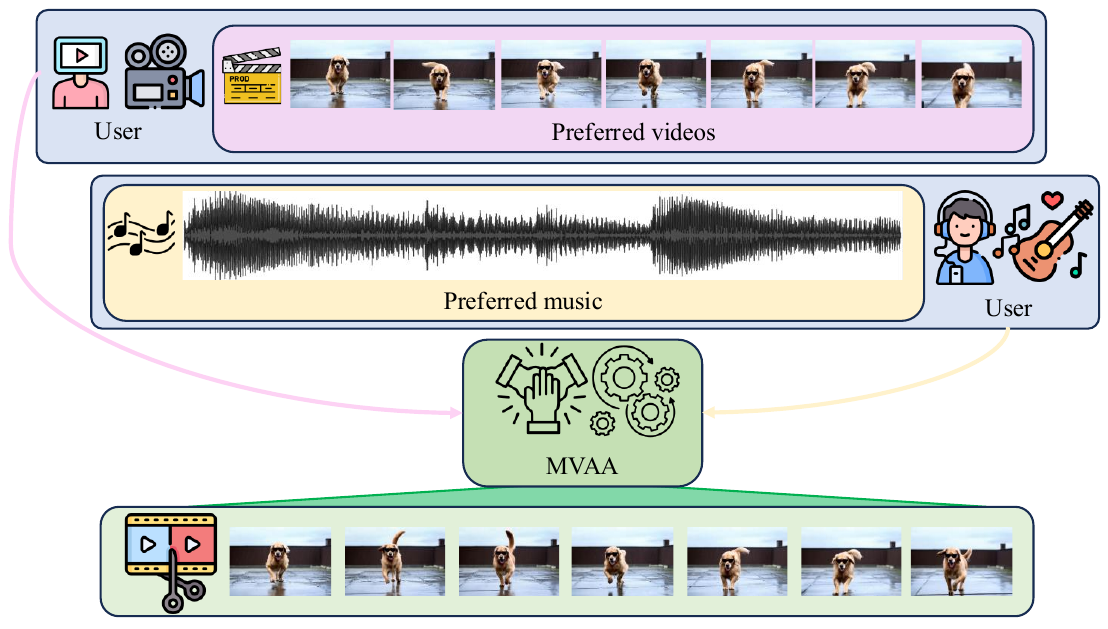}
  \caption{Our MVAA (Music-Video Auto-Alignment) is an automatic, efficient, and powerful music-driven video editing system to let the users' preferred videos to align any music what they like.}
  \label{fig:teaser}
\end{teaserfigure}

\maketitle

\section{Introduction}

Aligning the rhythm of motion in user-provided video footage with the beat structure of a chosen music track is a practical and valuable capability for filmmakers, advertisers, and social media creators. By aligning visual motion to musical beats, creators can enhance viewer engagement and narrative flow. However, current workflows depend on manual editing—including clip cutting, speed adjustments, and heuristic rule-based methods—that demand substantial time, domain expertise, and must be repeated whenever the music or footage changes. This labor-intensive process hinders rapid iteration, creative experimentation, and efficient content production.

Despite rapid progress in diffusion-based video synthesis, including text-to-video~\cite{singer2023makeavideo}, image-to-video~\cite{hong2023cogvideo,yang2024cogvideox}, and audio-driven editing~\cite{sung2023sound,li2024vidmusician}, these methods primarily focus on generating new content or conditioning freshly synthesized videos on audio. By contrast, music-driven video editing leverages existing raw clips and tracks, a capability that remains underexplored yet holds clear practical value. While some generative models handle joint video and music generation \cite{ruan2023mm,wang2024av}, they often entangle the two modalities, limiting flexibility in aligning video to music beats while preserving the full visual content.

In this paper, we explore the novel \textit{music-driven video editing} problem for aligning video motion to an arbitrary given music track with beat awareness, while preserving the video contents. To enable flexibility in the process, we propose a reformulation of music-driven video editing into two modularized and interdependent stages. In the first stage, we extract beat timestamps from the audio and detect salient motion peaks in the video, then establish sparse beat-motion correspondences by shifting keyframes to align with musical beats. In the second stage, we treat synchronization as a completion problem: a frame-conditioned diffusion model performs rhythm-aware inpainting to synthesize intermediate frames that seamlessly flow between the aligned anchors. These two operations enable explicit beat-aware alignment between video and music by adjusting video movements to match musical beats through controlled editing.

Our framework, 	Music-Video Auto-Alignment (MVAA), builds on this two-stage reformulation by addressing the fact that pretrained diffusion models alone cannot complete video segments with the fidelity required for seamless beat synchronization. To ensure high-quality inpainting of the to-be-edited video, we introduce test-time training, fine-tuning our auxiliary video completion model (AVM) on the target clip so it learns specific motion dynamics, textures, and lighting cues. This video-specific adaptation is crucial for preserving the original semantics while generating rhythm-aware frames, and it enables flexibility in alignment with arbitrary music tracks. Although full test-time training yields the best completion quality, it can be computationally intensive. To mitigate this, we propose a hybrid strategy: we first pretrain general motion priors on a small collection of videos, then perform rapid fine-tuning—just one epoch (approximately 10 minutes on a single NVIDIA GTX 4090 GPU)—to achieve near-optimal quality with substantially reduced runtime.

We evaluate MVAA on diverse video–music pairs using a quantitative beat-alignment metric, content preservation metric, and user studies to assess perceptual quality. Our results demonstrate that MVAA achieves high performance on rhythmic alignment and visual smoothness, without sacrificing the original content’s semantics.

\begin{figure}[!t]
  \centering
  \includegraphics[width=\linewidth]{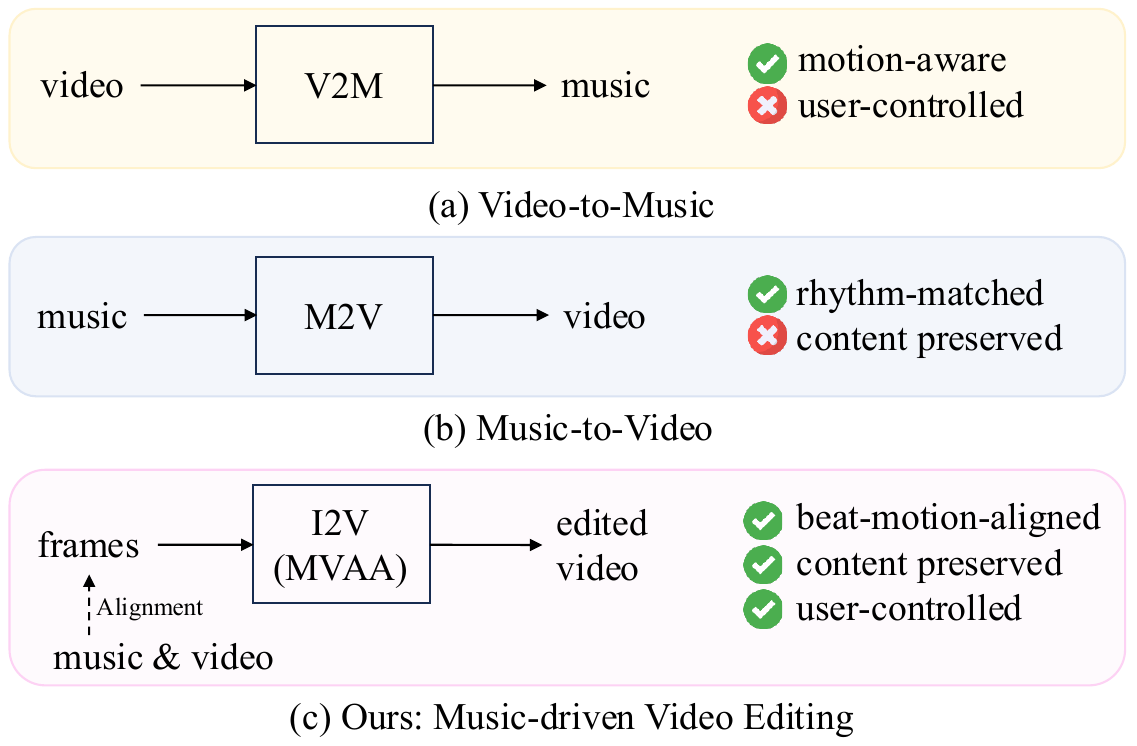}
  \caption{(a) Video-to-Music (V2M) models generate music conditioned on video content, but often lack controllability and produce synthetic music that may not match user preferences. (b) Music-to-Video (M2V) models generate video from music but struggle to preserve visual content and semantics. (c) Our approach, MVAA, performs music-driven video editing by automatically aligning existing video frames to music beats—ensuring rhythmic synchronization while preserving content and offering user control.}
\end{figure}

In summary, our contributions are:
\begin{itemize}
\item A clearly motivated definition of music-driven video editing, emphasizing its practical significance and current labor-intensive workflows.
\item A novel reformulation as beat-motion alignment followed by rhythm-aware inpainting, highlighting the conceptual elegance of our two-stage design.
\item The MVAA framework, which leverages sparse beat-aligned keyframes and a diffusion-based AVM trained via a hybrid pretraining and rapid fine-tuning strategy.
\item Comprehensive quantitative and user-study evaluations demonstrating the efficiency and effectiveness of our approach over existing methods.
\end{itemize}

\section{Related Work}

\subsection{Video Generation from Music and Audio}
Different from image generation and editing~\cite{ho2020denoising,betker2023improving,song2021scorebased,xie2025dymo,saharia2022photorealistic,nichol2021glide,ho2022classifierfree}, video generation can leverage audio as a conditioning input due to its temporal nature.
Early music-driven video generation approaches relied on retrieving and sequencing images based on lyrics \cite{musicstory,Cai2007AutomatedMV} or matching audio and visual content at perceptual and semantic levels \cite{Hua2004AutomaticMV,Shin2016AutomatedMV,emv-matchmaker,Fan2016DJMVPAA,liu2023emotionaware,Lin2017AutoMV}.
For general audio-driven video generation, Sound2Sight~\cite{chatterjee2020sound2sight} introduced a deep variational encoder-decoder that predicts future video frames conditioned on past frames and audio signals. TATS~\cite{ge2022long} tackled audio-to-video generation by combining a time-agnostic VQGAN with time-sensitive transformers to model temporal dynamics.
Recent advances in diffusion models \cite{ho2020denoising,song2020denoising} have enabled more expressive joint audio-video generation. For example, MM-Diffusion~\cite{ruan2023mm} allows bidirectional conditioning, enabling generation of video from audio and vice versa.
Some works specifically target music-conditioned human motion and dance video generation \cite{zhang2022music,zhuang2022music2dance,ofli2011learn2dance,alexanderson2023listen}, focusing on aligning motion dynamics with musical rhythm and beat.

\subsection{Music Generation with Conditions}
Music generation has been explored in both unconditional settings \cite{maina2023msanii,mittal2021symbolic,liu2020unconditional} and conditional contexts, where models are guided by various inputs. Among conditional approaches, text-to-music generation is a prominent direction, with models generating music conditioned on textual prompts \cite{yang2023diffsound,deng2024composerx,schneider2023mo}.
Beyond text, multi-modal conditioning has gained traction. Generative models are trained to generate music from visual conditions, including videos \cite{gan2020foley,hussain2023m,kang2023video2music,tian2024vidmuse}.  M$^2$UGen~\cite{hussain2023m} leverages large language models to process and generate music from combined video, audio, and text inputs. Video2Music~\cite{kang2023video2music} focuses on generating music that aligns with the visual content and emotional tone of a given video.
Recent works like V2Meow~\cite{su2023v2meow} and MeLFusion~\cite{chowdhury2024melfusion} enable music generation conditioned on video and image, respectively, with support for style control via text prompts. VidMuse~\cite{tian2024vidmuse} enhances alignment by incorporating both short-term and long-term modeling of visual features, enabling high-fidelity, rhythm-aware music generation from video inputs.

\subsection{Audio-Visual Alignment} 

Audio-visual alignment seeks to establish semantic or temporal correspondences between audio and visual modalities, playing a critical role in multi-modal understanding and generation tasks \cite{akbari2021vatt,rouditchenko2020avlnet,shi2022learning,cheng2022joint,gong2022contrastive,wu2023next,xing2024seeing}. Models such as CAV-MAE~\cite{gong2022contrastive} integrate contrastive learning with masked modeling to jointly learn audio-visual representations, while large-scale frameworks~\cite{girdhar2023imagebind} and multi-modal LLMs~\cite{wu2023next} extend alignment capabilities across diverse modalities, including audio and vision.

Beyond semantic alignment, recent works have addressed the synchronization of temporal dynamics between audio and video. This includes lip-speech alignment~\cite{halperin2019dynamic,chung2017out} and dance-music synchronization~\cite{bellini2018dance,yu2022self,zhou2023let}. AlignNet~\cite{AlignNet} learns cross-modal correspondences by mapping visual motion to audio rhythm using task-specific training. Some methods \cite{davis2018visual,chen2011visual,liao2015audeosynth,sun2023eventfulness} perform synchronization by extracting and aligning visual beats to musical beats through temporal warping. Specifically, VisBeat \cite{davis2018visual}, in particular, proposes a flow-based strategy that captures visual rhythm and aligns it with music via unfolding-based time-warping. Apart from heuristic, manually guided synchronization methods, existing approaches typically rely on task-specific training and are constrained by predefined scopes of video content and motion, or by the distributions seen during training. In contrast, our proposed task focuses on aligning video motion with music beats while preserving visual contents and generalizing to arbitrary music and video in an automated manner. The method supports test-time adaptation to a newly given video to further improve the performance.

\begin{figure}[t]
  \centering
  \includegraphics[width=\linewidth]{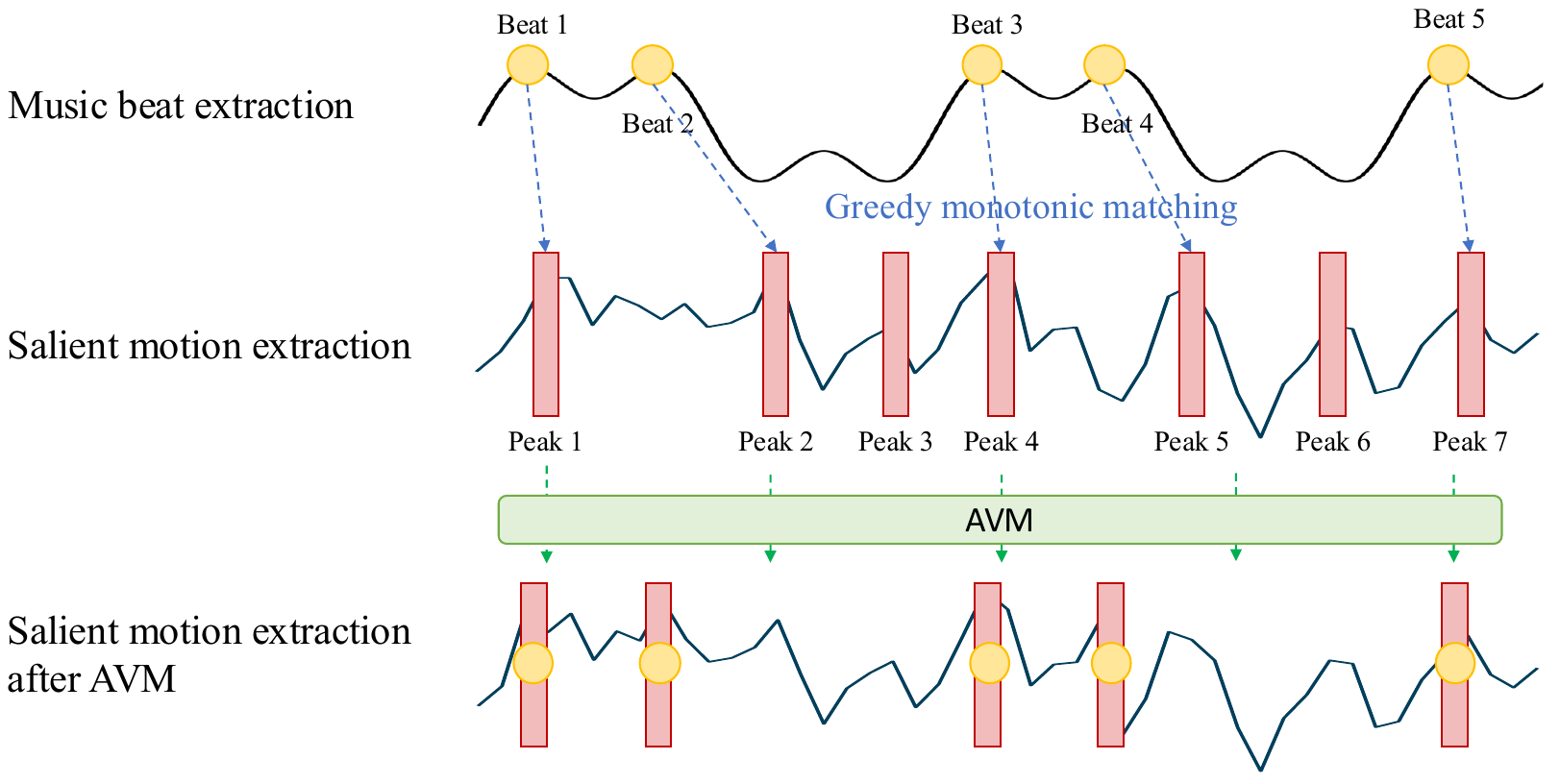}
  \caption{Overview of our Music-Video Auto-Alignment (MVAA) pipeline. We first perform beat-to-motion alignment by extracting music beats and salient motion peaks, followed by greedy monotonic matching to select beat-aligned keyframes from the input video. These keyframes are then used to guide the auxiliary video completion model, which inpaints the remaining frames to produce a temporally coherent and rhythmically aligned output video.}
  \label{fig:alignment}
\end{figure}

\section{Method: Music-Video Auto-Alignment}

\subsection{Task Definition}
Given an input video and a target music track, our objective is to automatically edit the video so that its visual rhythm aligns with the beat structure of the music. Unlike prior tasks such as music-to-video generation~\cite{li2024vidmusician, mittal2021symbolic,liu2020unconditional} or video-to-music composition~\cite{kang2023video2music,tian2024vidmuse}, we focus on a \textbf{novel} direction: enabling users to edit \textbf{any preferred video} to match the rhythmic structure of \textbf{any chosen music}. We refer to this task as \textbf{music-driven video editing}.

To formulate this task, we treat it as a generalized \textit{arbitrary-frame video completion} problem. Given a set of selected video frames $\{x_{t_1}, x_{t_2}, \dots, x_{t_k}\}$ in time indices $\{t_1, t_2, \dots, t_k\}$, our goal is to place them on new time indices $\{t^n_1, t^n_2, \dots, t^n_k\}$ and to synthesize the full video $\{x_1, x_2, \dots, x_L\}$ that is temporally smooth and rhythmically aligned with the inserted keyframes at new indices.

The core insight behind this formulation is that, by extracting a set of musical beat timestamps $\{b_1, b_2, \dots, b_k\}$ from the audio, we can select and relocate video frames to these corresponding timepoints. If a video completion model can robustly inpaint the missing frames while preserving semantic content and motion continuity, then we can achieve rhythm-aligned video editing by simply providing sparse, beat-guided frame anchors.

Importantly, this task formulation offers several advantages:
\begin{itemize}
  \item It allows flexible insertion of any number of keyframes at arbitrary temporal locations, enabling fine-grained control over rhythmic alignment.
  \item It operates in an editing paradigm rather than full video generation, ensuring that the original video content and identity are preserved.
  \item It is agnostic to the specific content of the video or music, allowing generalization across diverse inputs.
\end{itemize}

In this work, we present our \textit{Music-Video Auto-Alignment} (MVAA) framework. MVAA includes two modules, \ie, \textit{beat-to-motion alignment}, which is a lightweight preprocessing to extract the music beats and the corresponding video motion; and \textit{auxiliary video completion module}(AVM) to inpaint the remaining frames. In the following sections, we describe how we implement and build the whole model to support arbitrary beat-driven video editing.

\subsection{Beat-to-Motion Alignment}\label{sec:beat-to-motion-align}

To align the temporal rhythm of a video with a target music track, we identify and match rhythmic structures in both modalities. Specifically, we extract music beats from the audio and salient motion points from the video, and compute a one-to-one temporal alignment between them. This alignment serves as the basis for music-driven video editing.

\myparagraph{Music Beat Extraction.}
We extract beat times from the input music using a standard onset-based beat tracking algorithm. In particular, we apply \texttt{librosa}'s implementation of spectral flux onset detection followed by tempo estimation and beat tracking~\cite{ellis2007beat}. This yields a sequence of time-stamped beat locations $\mathcal{B} = \{b_1, b_2, \dots, b_M\}$ representing the rhythmic structure of the music.

\myparagraph{Salient Motion Extraction.}
To extract rhythmic cues from the video, we compute a motion energy signal by measuring frame-to-frame pixel-wise difference in the RGB space~\cite{virtanen2020scipy}. We then smooth this signal using a Gaussian filter and detect local maxima as motion peaks. These peaks, $\mathcal{V} = \{v_1, v_2, \dots, v_N\}$, correspond to time points where the visual activity is most salient, \eg, a step, jump, or gesture. These frames are later used as candidate anchors.

\myparagraph{Greedy Monotonic Matching.}
We adopt a simple yet effective one-to-one alignment strategy, named greedy monotonic matching: each music beat is matched to a unique motion peak, ensuring non-repetition and temporal ordering. Let $\mathcal{B} = \{b_1 < b_2 < \dots < b_M\}$ be the sorted beat timestamps, and let $\mathcal{V} = \{v_1 < v_2 < \dots < v_N\}$ be the sorted motion peaks.
We then compute a one-to-one monotonic matching $\pi: \mathcal{B} \rightarrow \mathcal{V}$ that minimizes the total alignment error:
\begin{equation}~\label{eq:gmm}
\pi^* = \arg\min_\pi \sum_{i=1}^K |b_i - v_{\pi(i)}|, \quad \text{subject to } \pi(i) < \pi(i+1),
\end{equation}
where $K$ is the $\texttt{min}(N, M)$.
This objective ensures that each selected motion anchor follows the unique musical beat structure while preserving chronological order. Figure~\ref{fig:alignment} illustrates an example of greedy monotonic matching overlaid on motion intensity curves.

\subsection{Auxiliary Video Completion Model}
To enable beat-synchronized video editing, we propose an auxiliary video completion model (AVM) that can synthesize temporally coherent video given a sparse set of arbitrary keyframes. Unlike standard image-to-video (I2V) diffusion models which take fixed start/end frames or uniformly spaced frames, our formulation generalizes to arbitrary conditioning frames placed at arbitrary time steps.

\myparagraph{Arbitrary Frame Conditioning.}
We \textbf{reformulate} the I2V task into an arbitrary inpainting problem: 
given a video length \( L \) and a subset of selected frames 
\(\{x_{t_1}, x_{t_2}, ..., x_{t_K}\}\), where \( t_k \in [1, L] \), 
our goal is to complete the full video sequence \( \mathbf{x}_{1:L} \) 
that is temporally smooth and consistent with the conditioning frames. These conditioning frames can be sampled at any timestep and do not need to be uniformly spaced or positioned at the start/end. This formulation allows flexible control over rhythmic synchronization, as we can insert keyframes at beat-aligned timestamps and let the model inpaint the rest.

\begin{figure}[t]
  \centering
  \includegraphics[width=0.95\linewidth]{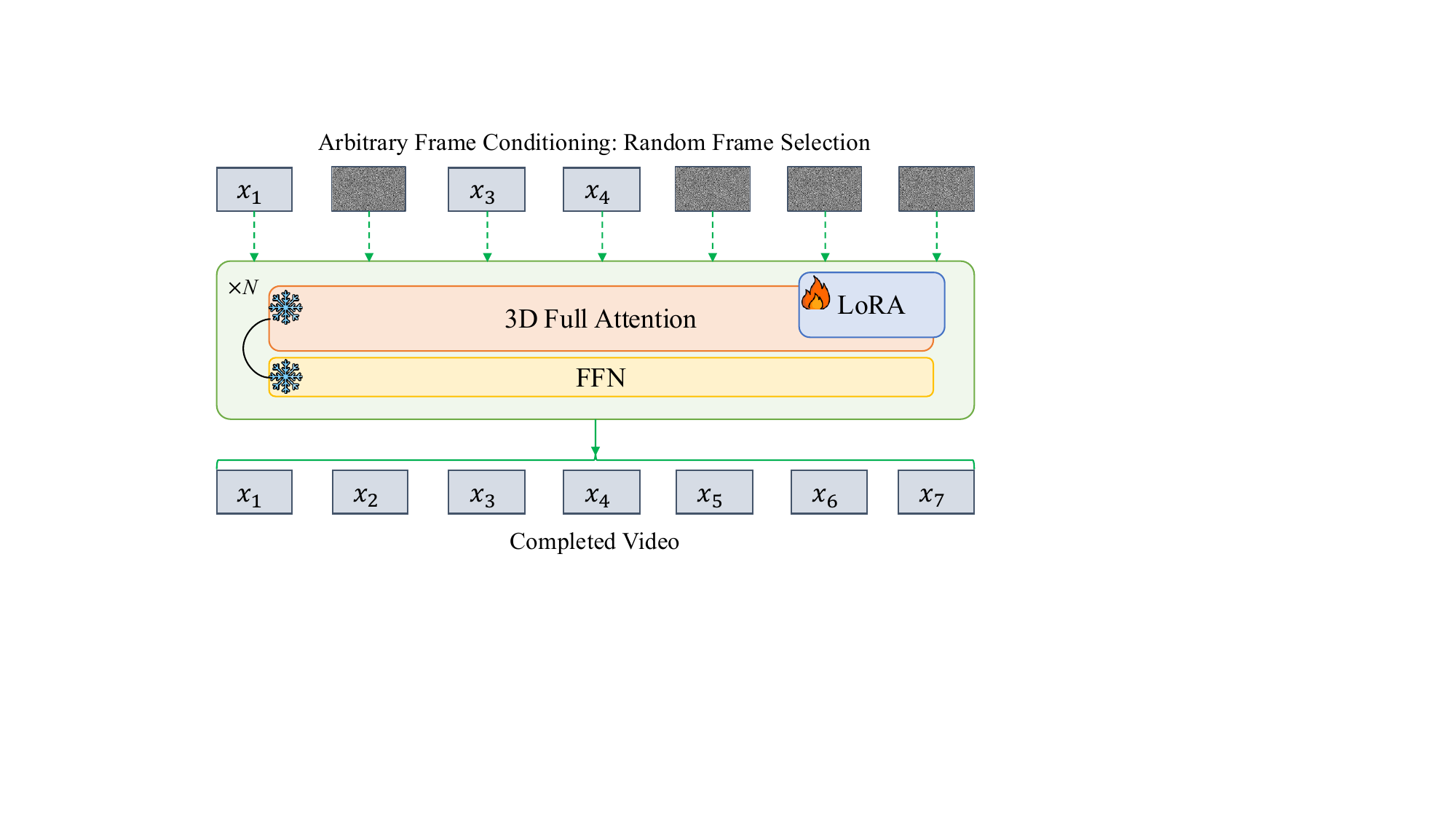}
  \caption{Overview of our auxiliary video completion model (AVM).}
  \label{fig:completion}
  \vspace{-0.2cm}
\end{figure}

\myparagraph{Model Training Strategy.} Our method builds upon conditional diffusion models~\cite{ho2020denoising, ho2022video}. Given a clean video sequence $\mathbf{x}_{1:L}$ of length $L$, the forward diffusion process progressively adds Gaussian noise to generate a noisy latent at step $t$:
\begin{equation}~\label{eq:add_noise}
q(\mathbf{x}^{(t)} | \mathbf{x}^{(0)}) = \mathcal{N}(\mathbf{x}^{(t)}; \sqrt{\bar{\alpha}_t} \mathbf{x}^{(0)}, (1 - \bar{\alpha}_t)\mathbf{I}),
\end{equation}
where $\mathbf{x}^{(t)}$ is the video latent with noise at timestep $t$.
During training, the model learns to predict the added noise $\epsilon$ using a neural network $\epsilon_\theta$. We follow the I2V model~\cite{yang2024cogvideox} to give a first frame $x_{1}$ from the video sequence as the condition. The model learning is formulated as:
\begin{equation}~\label{eq:loss}
    \mathcal{L}(\theta) := \mathbb{E}_{\mathbf{x}^{(0)}, y, \epsilon, t, x_{1}}{ \left\| \epsilon - \epsilon_\theta(\mathbf{x}^{(t)}, t, y, x_{1})
    \right\|^2},
\end{equation}
where $y$ is the condition, where is usually a text prompt.
To enable arbitrary video frame insertion at test time, we modify the training process by \textit{randomly masking out} a subset of frames in each training sample. Formally, for each training video $\mathbf{x}_{1:L}$, we randomly select a set of visible frames $\mathcal{V} = \{x_{t_k} \}_{k=1}^{K}$ with $t_k \in [1, L]$, and mask out the remaining frames. The model is then trained to reconstruct the full sequence conditioned only on the visible subset:
\begin{equation}~\label{eq:loss_modify}
\mathcal{L}_{\text{AVM}} = \mathbb{E}_{\mathbf{x}^{(0)}, y, \epsilon, t, \mathcal{V}} \left[ \left\| \epsilon - \epsilon_\theta(\mathbf{x}^{(t)}, t, y, \mathcal{V}) \right\|^2 \right].
\end{equation}
Rather than relying on fixed frame-beat pairs during training, we train the model using only video data, augmented by randomly sampling arbitrary conditioning frames at each iteration.
This strategy can teach the model to interpolate or extrapolate missing frames. Once trained, the model can accept keyframes placed at any timestamps—such as beat-aligned frames—and complete the video in a temporally coherent manner.

To improve efficiency and preserve general knowledge from the pretrained base model, we fine-tune the CogVideoX backbone using Low-Rank Adaptation (LoRA)~\cite{hu2021lora}. This allows us to inject temporal inpainting capability into the model by updating only a small number of parameters. Empirically, we find that training with LoRA on only one video is sufficient to adapt to new video content and yield high-quality completions.

\subsection{Test-time Training and Generalization} 
A unique advantage of our AVM is its adaptability across different video domains. While the model can be trained on a single input video, it demonstrates impressive generalization ability and can be further refined at test time to improve performance.

\myparagraph{Test-Time Fine-Tuning.} We first observe that even when trained on a single short video clip, the AVM can effectively reconstruct and align rhythmic content to arbitrary given music tracks. 
Moreover, when applying this single-video trained model to a new video, a short test-time fine-tuning phase allows rapid adaptation.
Specifically, fine-tuning the model on the new targeted video for just 1 epoch (approximately 10 minutes with 50 iterations on a single NVIDIA 4090) is sufficient to significantly enhance visual fidelity and beat synchronization.
This makes our approach highly practical for real-world applications with limited data and time constraints.

\myparagraph{Multi-Video Generalization.} To enhance the robustness and generalization ability of our AVM, we extend training to include multiple videos. This multi-video pretraining stage enables the model to generalize to diverse video content, allowing it to perform beat-aligned video completion without requiring any additional test-time tuning.
Empirically, we observe that training on as few as 10 videos already produces a strong general-purpose model capable of rhythm-aware editing. Scaling up to 1,000 diverse videos further improves generalization, enabling high-quality editing directly on previously unseen content without any further test-time fine-tuning.
Moreover, we find that pretraining with a larger dataset significantly accelerates test-time adaptation. For instance, test-time tuning on a specific video with the model pretrained with 1,000 videos—using only 1/1,000 of the total training iterations—achieves better performance than directly training from scratch on this video.

\myparagraph{Balancing Generalization and Adaptation.}
While training on multiple videos improves zero-shot generalization, we find that optional test-time fine-tuning on the targeted video can further bring additional performance gains. 
This flexibility enables users to either deploy a general model or optionally specialize it for a specific video domain with minimal additional cost.

\begin{figure*}[t]
  \centering
  \includegraphics[width=0.92\linewidth]{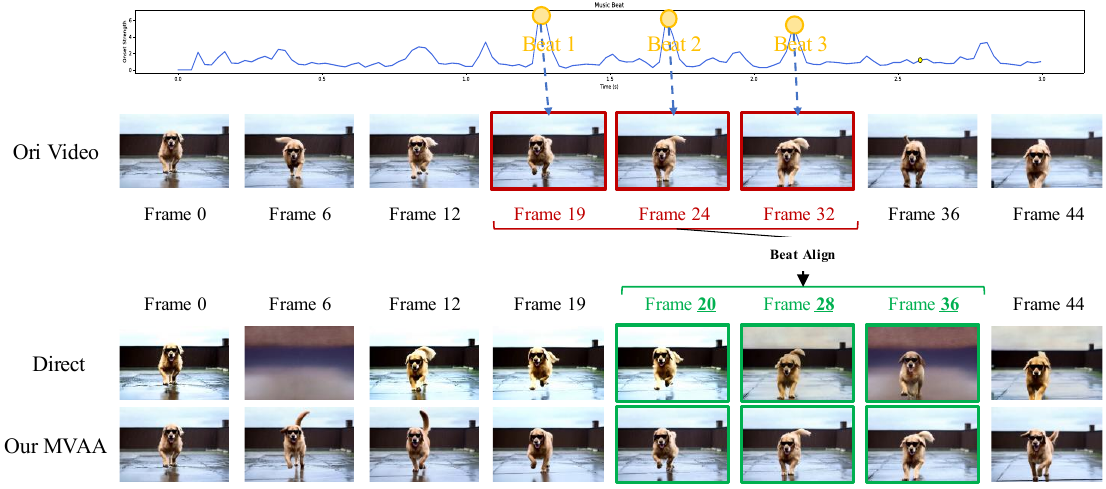}
  \caption{Qualitative comparison of music-driven video editing. Our method (MVAA) exhibits stronger motion-beat synchronization compared to baselines. Keyframes extracted from the original video (in red) are aligned with the correct music beat timestamps (in green).}
  \label{fig:qualitative}
\end{figure*}
\section{Experiments}

\subsection{Datasets}

\myparagraph{Training Data.}
We conduct experiments on both single-video and multi-video setups. For single-video settings, we directly use the test video as the training data to test how well our model can align a given video to arbitrary music. For multi-video training and generalization, we randomly sample a subset of $N=1000$ videos from OpenVid-HD~\cite{nan2024openvid} and VideoUFO~\cite{wang2025videoufo}, covering diverse motion styles and visual appearances.

\myparagraph{Benchmark Construction.} To the best of our knowledge, no existing benchmark directly targets music-driven video editing. We therefore construct a new evaluation suite comprising:

\textit{Video Clips.} We collect 10 video clips sourced from several recent state-of-the-art text-to-video generation platforms such as \textit{Veo 3}~\cite{veo3}, \textit{Sora}~\cite{sora}, \textit{Wan 2.1}~\cite{wan2025}, \textit{CogVideoX}~\cite{yang2024cogvideox}, and \textit{Imagine Art}~\cite{ImagineArt}. These videos cover a wide range of motion patterns and semantics, including human walking, dancing, object movement, and animal locomotion.

\textit{Music Tracks.} We assemble 5 music segments (3 seconds each), including Rocket Man, Bad Guy, Jingle Bells, Birds of a Father and a Chinese song Common Jasmine Orange. Beat annotations are obtained via standard onset-based beat tracking using \texttt{librosa}.
Each video can align with each music. In total, we have 50 music-video pairs. 

\subsection{Implementation Details}

\myparagraph{Training Configurations.}
We evaluate three training configurations to assess the flexibility and generalization capabilities of our model:

\textit{Single-video training.} We train the AVM using only one video clip. This setting simulates the practical use case of personalized, user-driven video editing, where only a single target video is available.

\textit{Multi-video training.} To enhance generalization, we train the model on a larger collection of videos. Specifically, we experiment with datasets containing 10 or 100 or 1,000 clips, demonstrating the scalability of our approach and its ability to handle diverse visual content and motion patterns.

\textit{Test-time adaptation.} After training on either a single video or a multi-video dataset, the model can be directly applied to new videos. In addition, we optionally apply test-time fine-tuning to further adapt the model to unseen video content. This process is lightweight—requiring only a few minutes of training—and consistently improves video quality and temporal coherence.

\myparagraph{Evaluation Metrics.}
We use the following automatic metrics to assess performance:

\begin{itemize}

\item \textit{Beat Alignment Score (BeatAlign):} BeatAlign~\cite{li2021ai} measures the synchronization between motion and music by quantifying the alignment between motion peaks and music beats in the generated videos. Both motion peaks and music beats are extracted as described in Section~\ref{sec:beat-to-motion-align}. The score is defined as the mean absolute error between the timestamps of motion-aligned frames and their corresponding music beat positions. Higher scores mean higher music-video alignment.
\item \textit{Temporal Consistency (TC):} We assess temporal coherence by computing similarity scores between consecutive frames, using features extracted from the CLIP-Large model~\cite{radford2021learning}. Higher scores indicate stronger temporal consistency across frames.

\item \textit{Content Preservation (CP):} We use learned perceptual image patch similarity (LPIPS) metric~\cite{zhang2018unreasonable} quantify perceptual similarity and content preservation. Lower is better.
\end{itemize}

\myparagraph{User Study.} 
To assess the perceptual effectiveness of our beat-aligned video editing, we conduct a user study using pairwise comparisons between our method and strong baselines.

We use 50 music-video pairs from our test set and generate edited videos using various approaches. For each pair, we present two versions of the same video with synchronized music—one generated by our method and the other by a baseline—and ask participants following questions:

\begin{itemize}
\item \textit{Music-Video Alignment Quality (MVA): }``Which video better aligns visual motion with the music beats?''
\item \textit{Overall Preference (All): }``Which version do you prefer overall, including the music-video alignment quality, video smoothness and visual quality.
\end{itemize}

Each participant compares 50 pairs (randomized order and left/right position). We collect responses from 20 participants, resulting in 1000 pairwise judgments in total. We compute the \textbf{win rate}—the percentage of times our method is preferred over each baseline.

\myparagraph{Model Details.}
We use CogVideoX-5B-I2V~\cite{yang2024cogvideox} as our diffusion backbone, fine-tuned using LoRA~\cite{hu2021lora} with rank 64 and learning rate $1\mathrm{e}{-5}$. We train 10,000 iterations for the single-video training setting and the 10-video training setting, and 70,000 iterations for the other multi-video training settings. All videos are resized to $480{\times}720$ and processed at 16 fps, following \cite{yang2024cogvideox}. Inference is performed by inserting beat-aligned keyframes and completing the sequence using AVM.

\subsection{Main Results}

We first evaluate whether our model can effectively align visual motion with musical beats. To this end, we compare MVAA against the following baselines: 1) the original videos, and 2) direct application of the CogVideoX-5B-I2V model~\cite{yang2024cogvideox} for arbitrary frame interpolation. As shown in Table~\ref{tab:main_results}, our method achieves strong performance on the novel task of music-driven video editing and obtains high music-beat alignment scores.

Figure~\ref{fig:qualitative} presents qualitative comparisons demonstrating that baseline methods struggle with this task. In particular, CogVideoX \cite{yang2024cogvideox} produces flickering artifacts due to its lack of support for arbitrary interpolation. In contrast, MVAA generates smoother, rhythmically aligned, and semantically coherent video sequences, highlighting its effectiveness for music-synchronized video synthesis.

\begin{table}[t]
  \centering
  \caption{Quantitative comparison on music-driven video editing. ``Direct'' represents directly using CogVideoX-5B-I2V\cite{yang2024cogvideox} model for arbitrary video completion.}
  \label{tab:main_results}
  \begin{tabular}{lccc}
    \toprule
    Method & BeatAlign $\uparrow$ & TC $\uparrow$ & LPIPS $\downarrow$ \\
    \midrule
    Original Video & 0.204 & \textcolor{gray}{0.963} & \textcolor{gray}{-} \\
    Direct~\cite{yang2024cogvideox} & 0.273 & 0.835 & 0.001 \\
    \textbf{MVAA (Ours)} & \textbf{0.312} & \textbf{0.949} & \textbf{0.000} \\
    \bottomrule
  \end{tabular}
  \vspace{-0.5cm}
\end{table}

\myparagraph{Long-video, long-music alignment.}
Our MVAA is capable of generating long videos that align with extended music tracks. In this experiment, we generate the final output by sequentially producing short video clips and concatenating them. Only the first short clip is adapted using test-time training; all subsequent clips are edited directly without additional adaptation.

\begin{figure}[t]
  \centering
  \includegraphics[width=\linewidth]{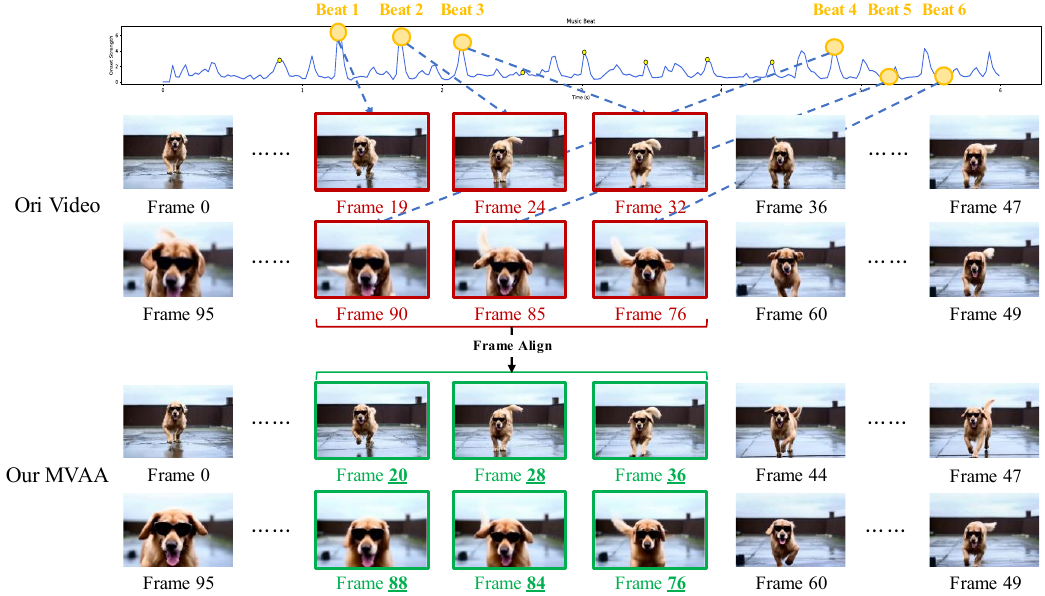}
  \caption{Visualization of long-video, long-music alignment.}
  \label{fig:longvideo}
\end{figure}

As shown in Figure~\ref{fig:longvideo}, MVAA can successfully produce long videos using just a single short clip for adaptation, demonstrating strong generalization capabilities. Moreover, MVAA maintains long-term content consistency across the concatenated segments.

It is worth noting that our current approach to long-video generation, \ie, naively concatenating short clips, is a simple and challenging strategy. In future work, we aim to incorporate advanced long-video generation techniques, such as~\cite{wu2024moviebench,lufreelong,zhao2025riflex,wang2024framer}, to further enhance long-video quality.

\subsection{Ablation Study}

\myparagraph{Effect of Pretraining Video Numbers.}
We explore how the number of pretraining videos affects final performance. We directly evaluate the pre-trained model on the benchmark.
As shown in Table~\ref{tab:num_videos}, increasing the number of training clips consistently improves BeatAlign and MVA, indicating that more pretraining data enhances the model’s ability for arbitrary video completion.
The slight decrease in TC and LPIPS can be attributed to underfitting in motion smoothness, as the number of training iterations increased by only 7$\times$ despite a 10$\times$ to 100$\times$ increase in video data.
Nonetheless, the user study reveals that additional video pretraining significantly boosts overall quality, demonstrating superior visual fidelity, generalization, and alignment between music and video.

\myparagraph{Effect of Test-Time Training.} We further evaluate whether test-time adaptation can improve performance across models pretrained on datasets of varying sizes. In this experiment, we conduct both quantitative evaluation and user studies, comparing models that share the same pretraining data—with and without test-time tuning.

As shown in Table~\ref{tab:testtime}, test-time tuning consistently improves the BeatAlign metric, demonstrating its effectiveness in enhancing rhythm synchronization. We highlight the following observations:
1) Regardless of the pretraining dataset size, test-time tuning consistently improves the BeatAlign score. This underscores the utility of test-time adaptation in helping the model more precisely align motion with musical beats.
2) While test-time tuning improves BeatAlign, it slightly degrades TC and LPIPS scores under the 10-video setting. We attribute this to the limited pretraining data, which causes test-time tuning to compensate for insufficient learning, ultimately leading to poorer overall video quality despite improved beat-conditioned completion.
3) Interestingly, applying test-time tuning on a single video pretraining (for the same number of iterations as the 10-video setting) yields even better BeatAlign performance than pre-training on 10 videos. This suggests that overfitting a single video can still effectively transfer beat-aware completion to unseen content. However, this comes at the cost of lower temporal consistency, likely due to frame discontinuities.

To balance computational cost and performance, we use 1,000 videos in our current setup. We leave large-scale training for future work.

\begin{table}[t]
\centering
\setlength{\tabcolsep}{4pt} %
\renewcommand{\arraystretch}{1.0} %
\caption{Effect of the number of pretraining videos. Results of MVA and All are the win rates of pretraining on 1000 videos.}
  \label{tab:num_videos}
    \scalebox{1.0}{
      \begin{tabular}{lccc|cc}
        \toprule
        \# Pretrain & BeatAlign $\uparrow$ & TC $\uparrow$  & LPIPS $\downarrow$ & MVA $\uparrow$  & All $\uparrow$\\
        \midrule
        10            & 0.251 & \textbf{0.954} & 0.003 & 0.86 & 0.88 \\
        100           & 0.291 & 0.948 & \textbf{0.003} & 0.64 & 0.66 \\
        1000          & \textbf{0.304} & 0.947 & 0.006 & - & - \\
        \bottomrule
      \end{tabular}
    }
\end{table}

\myparagraph{Effect of Pretraining Video Quality.} To understand the impact of video quality on pretraining, we compare models trained on different sources: OpenVid-HD~\cite{nan2024openvid} and VideoUFO~\cite{wang2025videoufo}. 
As shown in Table~\ref{tab:quality}, training on OpenVid-HD yields consistently better results.
We put analysis on the video quality, we found that the video quality of OpenVid-HD is higher than that of VideoUFO, while VideoUFO having diverse categories. 

\begin{table}[t]
  \centering
  \caption{Impact of test-time adaptation. Results of MVA and All are the win rates of using test-time fine-tuning on the targeted videos.}
    \setlength{\tabcolsep}{2pt} %
    \renewcommand{\arraystretch}{1.0} %
      \label{tab:testtime}
    \scalebox{0.95}{
  \begin{tabular}{l|ccc|cc}
    \toprule
    Setup & BeatAlign $\uparrow$ & TC $\uparrow$ & LPIPS $\downarrow$ & MVA $\uparrow$  & All $\uparrow$ \\
    \midrule
    \textcolor{gray}{1 Video (+ test-time) }   & \textcolor{gray}{0.277} & \textcolor{gray}{0.930} & \textcolor{gray}{0.000} & - & - \\
    10 Video (pretrain)& 0.251 & \textbf{0.954} & 0.003 & 0.64 & 0.58  \\
    10 Video (+ test-time)& 0.283 & 0.950 & 0.006 & - & -  \\
    1000 Videos (pretrain) & 0.304 & 0.947 & 0.006 & 0.62 & 0.60 \\
    1000 Videos (+ test-time)  &  \textbf{0.312} & 0.949 & \textbf{0.000} & - & - \\
    \bottomrule
  \end{tabular}
  }
\end{table}

\begin{table}[t]
  \centering
  \caption{Impact of dataset quality on editing performance.}
  \label{tab:quality}
  \begin{tabular}{lccc}
    \toprule
    Dataset & BeatAlign $\uparrow$ & TC $\uparrow$ & LPIPS $\downarrow$ \\
    \midrule
    VideoUFO~\cite{wang2025videoufo}   & 0.297 & 0.945 & 0.006 \\
    OpenVid~\cite{nan2024openvid}    & \textbf{0.312} & \textbf{0.949} & \textbf{0.000} \\
    \bottomrule
  \end{tabular}
\end{table}

\begin{figure*}[t]
  \centering
  \includegraphics[width=\linewidth]{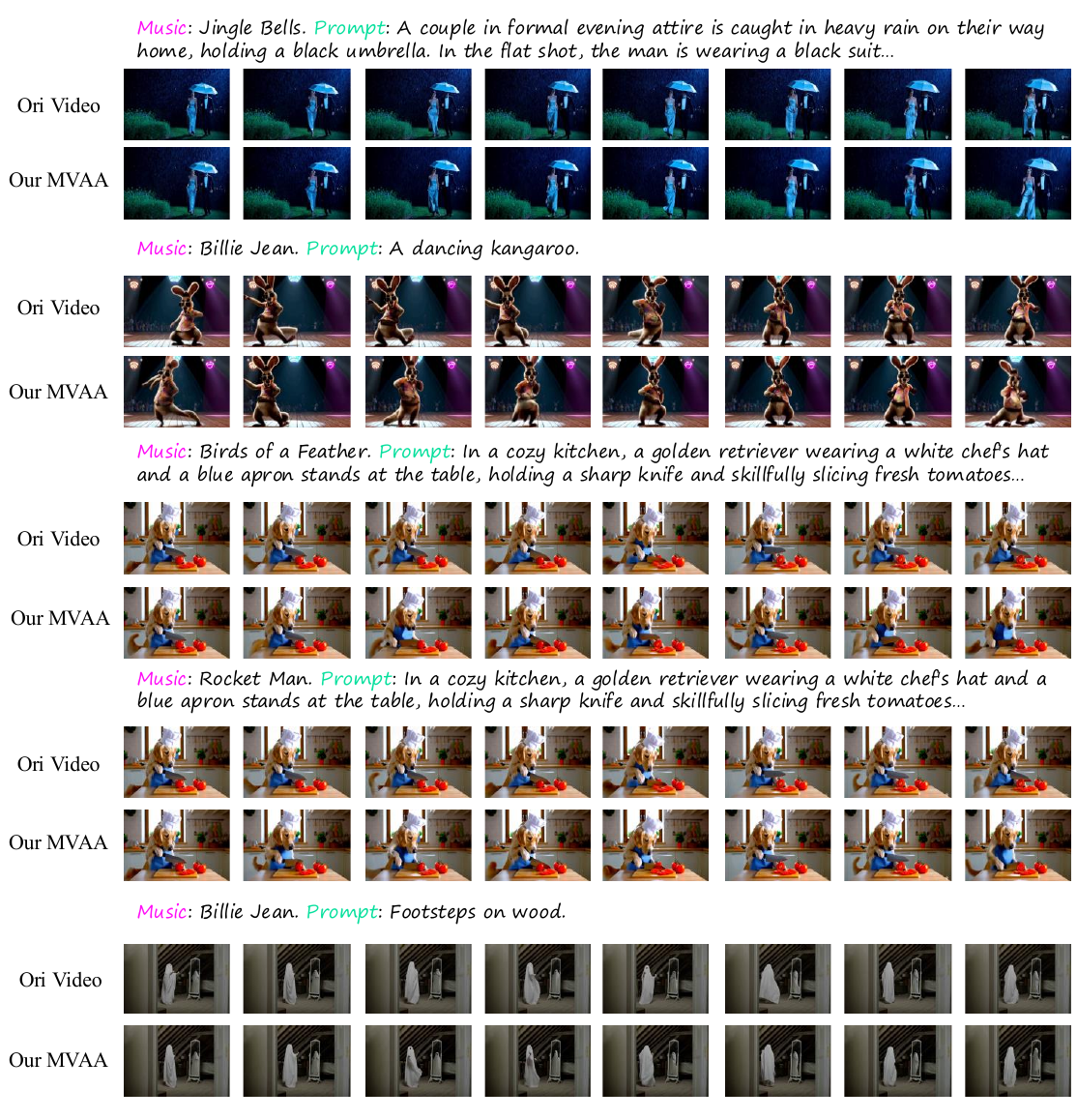}
  \caption{Visualization of beat-aligned video generation. Top: input video; bottom: output edited by our MVAA. Please refer to the project page for clearer visualization and music.
  }
  \label{fig:qual-compare}
\end{figure*}

\myparagraph{Qualitative Evaluation.}
We present qualitative comparisons in Figure~\ref{fig:qual-compare}, showing that MVAA produces smoother transitions and more accurately matches beat timing. 
Meanwhile, our MVAA can edit arbitrary videos to align beats from any given music.
The video samples are provided in the project page for better visualization.

\section{Conclusion}

We present a novel and flexible framework for synchronizing video motion to music beats through a two-step editing process: keyframe alignment and rhythm-aware video inpainting. By combining beat-conditioned keyframe insertion with a frame-conditioned diffusion model, our approach preserves visual content while achieving smooth, beat-aligned transitions. The hybrid training strategy, with lightweight fine-tuning, enables practical and efficient adaptation to specific videos. Experimental results and user studies confirm the effectiveness of our method in enhancing rhythmic coherence and visual appeal over baselines. 
In future work, we aim to extend this framework to handle more general audio sources beyond music, apply it to longer-form content such as films, and explore joint retrieval of suitable music for a given video to support end-to-end audiovisual editing.

\bibliographystyle{ACM-Reference-Format}
\bibliography{main}

\end{document}